\UseRawInputEncoding
\documentclass[pdflatex,sn-nature]{sn-jnl}% Math and Physical Sciences Numbered Reference Style
\usepackage[utf8]{inputenc}
\usepackage{graphicx}%
\usepackage{newunicodechar}
\newunicodechar{≥}{\geq}
\newunicodechar{−}{\textminus}
\usepackage{multirow}%
\usepackage{amsmath,amssymb,amsfonts}%
\usepackage{amsthm}%
\usepackage{mathrsfs}%
\usepackage[title]{appendix}%
\usepackage{xcolor}%
\usepackage{soul}
\usepackage{textcomp}%
\usepackage{manyfoot}%
\usepackage{booktabs}%
\usepackage{caption}
\usepackage{algorithm}%
\usepackage{algorithmicx}%
\usepackage{algpseudocode}%
\usepackage{listings}%
\usepackage{booktabs}
\usepackage{xcolor}
\usepackage{tikz}
\usepackage{pgfplots}
\pgfplotsset{compat=1.18}
\usepackage{geometry}
\usepackage{soulutf8} 
\sethlcolor{yellow} 

\theoremstyle{thmstyleone}%
\theoremstyle{thmstyletwo}%

\theoremstyle{thmstylethree}%
\begin{document}

\title[Article Title]{The Alignment Paradox of Medical Large Language Models in Infertility Care: Decoupling Algorithmic Improvement from Clinical Decision-making Quality}


\author[1,2,6]{\fnm{Dou} \sur{Liu}}
\equalcont{These authors equally contributed to this work.}
\author[1,6,7]{\fnm{Ying} \sur{Long}}
\equalcont{These authors equally contributed to this work.}
\author[3]{\fnm{Sophia} \sur{Zuoqiu}}
\author[3]{\fnm{Kaipeng} \sur{Xie}}
\author[3]{\fnm{Runze} \sur{Yang}}
\author[3,4,5]{\fnm{Di} \sur{Liu}}
\author[4,5]{\fnm{Kang} \sur{Li}}
\author[8]{\fnm{Yiting} \sur{Lin}}
\author[8]{\fnm{Hanyi} \sur{Liu}}

\author*[3,4,5]{\fnm{Rong} \sur{Yin}}
\email{rong.yin@scupi.cn}

\author*[1,6,7]{\fnm{Tian} \sur{Tang}}
\email{tiantang2016@scu.edu.cn}

\affil[1]{\orgdiv{Department of Obstetrics and Gynecology}, \orgname{West China Second University Hospital}, \orgaddress{\country{China}}}

\affil[2]{\orgdiv{Department of Industrial and Operations Engineering}, \orgname{University of Michigan},  \country{U.S.}}

\affil[3]{\orgdiv{Department of Industrial Engineering}, \orgname{Sichuan University}, \orgaddress{\country{China}}}

\affil[4]{\orgdiv{West China Biomedical Big Data Center}, \orgname{West China Hospital}, \orgaddress{\country{China}}}

\affil[5]{\orgdiv{Med-X Center for Informatics}, \orgname{Sichuan University}, \orgaddress{\country{China}}}

\affil[6]{\orgdiv{Key Laboratory of Birth Defects and Related Diseases of Women and Children},  \orgname{Sichuan University}, \orgaddress{\country{China}}}

\affil[7]{\orgdiv{Reproductive Medical Center, Department of Obstetrics and Gynecology, West China Second University Hospital}, \orgname{Sichuan University}, \orgaddress{\country{China}}}

\affil[8]{\orgdiv{West China School of Medicine}, \orgname{Sichuan University}, \orgaddress{\country{China}}}


\abstract{ 
Large language models (LLMs) are increasingly adopted in clinical decision support, yet aligning them with the multifaceted reasoning pathways of real-world medicine remains a major challenge. Using more than 8,000 infertility treatment records, we systematically evaluate four alignment strategies: Supervised Fine-Tuning (SFT), Direct Preference Optimization (DPO), Group Relative Policy Optimization (GRPO), and In-Context Learning (ICL) through a dual-layer framework combining automatic benchmarks with blinded doctor-in-the-loop assessments. GRPO achieves the highest algorithmic accuracy across multiple decision layers, confirming the value of reinforcement-based optimization for structured prediction tasks. However, clinicians consistently prefer the SFT model, citing clearer reasoning processes (p = 0.035) and higher therapeutic feasibility (p = 0.019). In blinded pairwise comparisons, SFT attains the highest winning rate (51.2\%), outperforming both GRPO (26.2\%) and even physicians’ original decisions (22.7\%). These results reveal an alignment paradox: algorithmic improvements do not necessarily translate into higher clinical trust, and can diverge from human-centered preferences. Our findings highlight the need for alignment strategies that prioritize clinically interpretable and practically feasible reasoning, rather than solely optimizing decision-level accuracy.
}




\maketitle

 \newpage  
\section{Introduction}\label{sec1}
Large language models (LLMs) have rapidly advanced across domains, yet how their post-training alignment interacts with the hierarchical, high-stakes reasoning of real-world medicine remains poorly understood. These alignment methods usually involve various reinforcement learning variants such as Reinforcement Learning with Human Feedback (RLHF), Group Relative Policy Optimization (GRPO), and Direct Preference Optimization (DPO), which have shown remarkable improvement within some general areas by using verifiable rewards or preference signals to optimize the policy model.  \cite{deepseek-ai_deepseek-r1_2025,rafailov_direct_2023,lai_med-r1_2025,shao_deepseekmath_2024,dao_alphamaze_2025}For example, DeepSeek-R1 adopted the GRPO and its performance on math questions significantly outperformed the baseline. However, clinical decision-making presents a fundamentally different challenge: reasoning chains are long and multi-dimensional, rewards are noisy or unverifiable, and clinicians rely heavily on explanation quality and feasibility rather than solely output correctness alone. These properties suggest a structural tension between \textit{algorithmic alignment} and \textit{clinical alignment}, a tension that has not been mechanistically examined.
The domain-specific medical LLMs, like Med-Palm, MedGemma, and Lingshu, were proven to exhibit profound potential in solving clinical problems, ranging from text classification to clinical image reports\\\cite{chen_huatuogpt-o1_2024,singhal_toward_2025,team_lingshu_2025,gaber_evaluating_2025}. These models demonstrate capabilities that surpass human-level performance in the answer correctness on benchmark datasets. Despite their outstanding performance, when applying these state-of-the-art (SOTA) models to specialized real-world clinical cases instead of a general consultation, which could support the realistic diagnosis operation, they seem to encounter some obstacles\cite{yang_large_2023,hager_evaluation_2024}. Although SOTA models generated responses were preferred to those provided by generalist physicians, they still did not outperform domain-specific specialists, indicating limited utility in addressing highly specialized and practical clinical problems \cite{liu_medchain_2025}. Moreover, their opaque reasoning processes limit clinical adoption, despite evidence that explanation-based systems (“XAI”) enhance trust and interpretability \cite{ullah_challenges_2024, dwivedi_explainable_2023, evans_explainability_2022, martinez-aguero_interpretable_2022}.  While supervised fine-tuning (SFT) remains the dominant paradigm in medical model development\cite{yang_large_2023,wei_finetuned_2022}, data scarcity and incomplete supervision motivate increasing reliance on post-training alignment\cite{wang_critique_2025,team_lingshu_2025,dai_qoq-med_2025}. Yet, despite the rapid adoption of RLHF-style optimization, little is known about whether improvements in algorithmic metrics translate to enhanced interpretability, trust, or multi-step clinical reasoning. Even fewer studies have evaluated how different alignment paradigms behave in real-world, interdependent clinical decision environments where reasoning quality directly affects patient safety. This gap raises a fundamental scientific question: \textbf{Does stronger post-training alignment produce more clinically aligned models—or can optimization distort clinical reasoning, giving rise to an alignment paradox?}

As one of the medical diseases and global health issues declared by the World Health Organization (WHO), Infertility is estimated to be experienced by one in six at some stage in their lives globally\cite{organization_infertility_2023}. Assisted Reproductive Technology (ART) has become a central clinical pathway for these patients, typically after the failure of conventional treatments such as cycle regulation or ovulation induction\cite{graham_assisted_2023}.   ART serves as an ideal testbed for reasoning alignment because it demands the rigorous integration of high-dimensional data, including age, Anti-Müllerian Hormone (AMH), Follicle-Stimulating Hormone (FSH), Body Mass Index (BMI), endocrine profiles, past medical history, and gynecological ultrasound findings, to formulate a safe and effective treatment plan. Beyond selecting the ART strategy, clinicians must also determine the controlled ovarian stimulation (COS) protocol and the gonadotropin (Gn) starting dose, decisions that are interdependent and sensitive to subtle variations in the patient’s physiological state.  Consequently, ART decision-making is a multi-stage, high-dimensional, and evidence-driven process that is both time-consuming and cognitively demanding. Clinical outcomes are highly dependent on nuanced reasoning: inappropriate treatment selection can reduce cycle success rates, increase financial and emotional burden, and even elevate the risk of severe complications such as ovarian hyperstimulation syndrome (OHSS)\cite{kumar_ovarian_2011}. These decisions rely heavily on individual clinical experience, leading to substantial inter-physician and inter-center variability and limiting the scalability of standardized training for novice practitioners. Moreover, high outpatient volumes can induce decision fatigue among experienced clinicians\cite{zheng_decision_2020}. The inherent complexity, multi-layered structure, and safety-critical nature of ART therefore make it not only a representative real-world medical decision environment, but also an exceptionally sensitive setting in which failures of model alignment, if present, are likely to be amplified and clinically consequential.

We therefore pose three key questions: 
    \begin{itemize}
        \item What is the performance of standard SFT when applied to real-world, multi-dimensional clinical decision-making?
        \item Can advanced post-training alignment strategies achieve superior performance on objective decision-making metrics in such settings, and what patterns of improvement or degradation emerge across decision layers?
        \item Critically, do gains in algorithmic alignment translate into higher clinical trust, or do they instead reveal a divergence between accuracy-based optimization and clinicians’ subjective assessments of reasoning quality and therapeutic feasibility?
    \end{itemize}

To address these questions, we systematically compare four representative alignment paradigms: SFT, DPO, GRPO, and In-Context Learning (ICL), using a large-scale real-world dataset of more than 8,000 infertility treatment records. To enhance robustness on challenging and long-tail cases, we construct a hierarchical “pyramid” dataset that emphasizes diverse difficulty levels and decision structures. Most importantly, we introduce a dual-evaluation framework that integrates automated metric-based assessment with triple-blind doctor-in-the-loop evaluations, enabling us to directly examine the relationship between algorithmic optimization and clinician-centered preferences. This framework allows us to uncover not only performance differences across alignment strategies, but also the underlying tension between algorithmic alignment and clinical alignment, a tension that forms the core of the alignment paradox we investigate in this study.

\begin{figure}
      \centering
\label{fig:Overview}
\caption{Overview of the Alignment and Evaluation Framework}
    \includegraphics[width=1\linewidth]{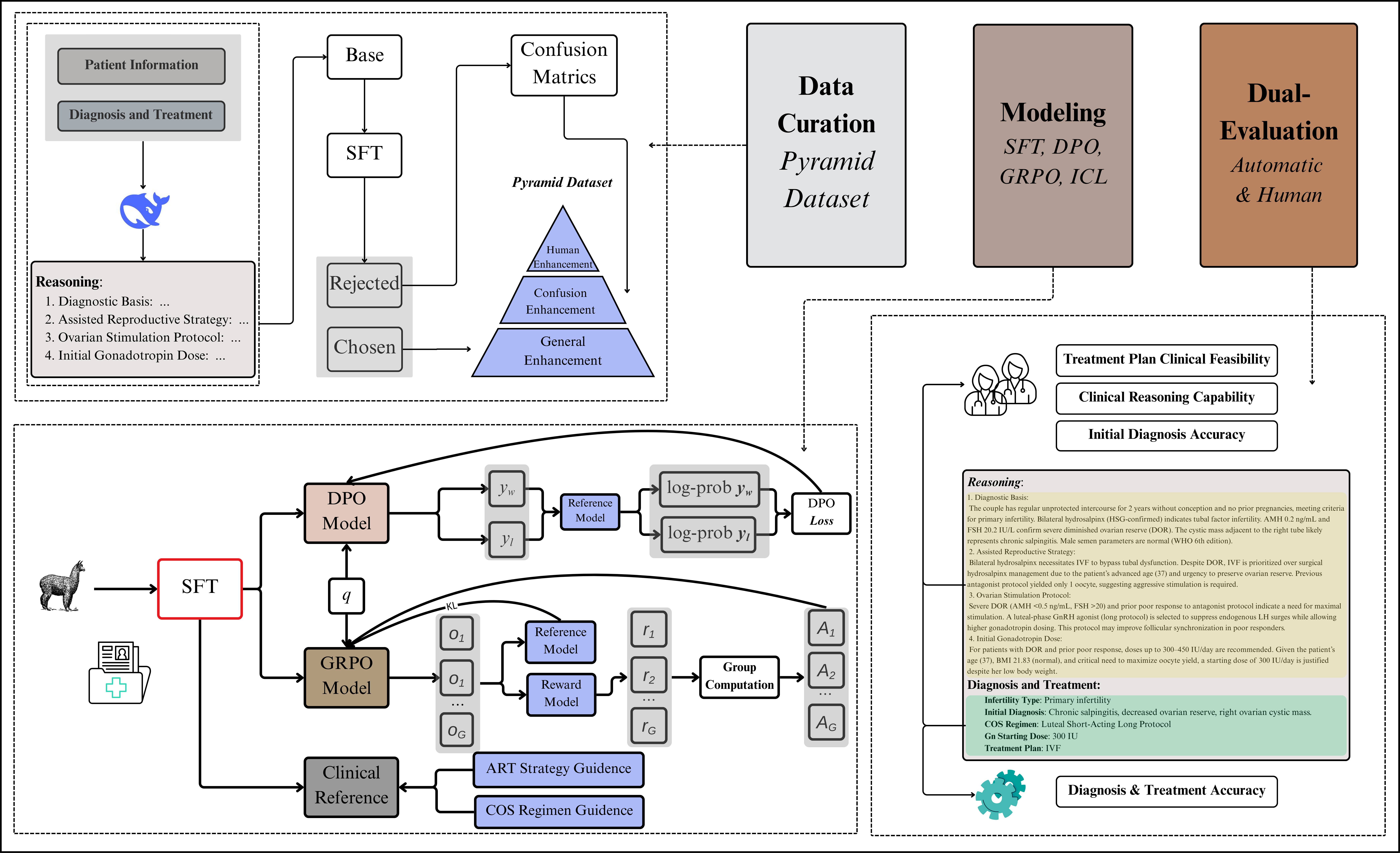}

\begin{flushleft}
The proposed dual-alignment framework integrates supervised fine-tuning (SFT) with post-training optimization methods, DPO, GRPO, and ICL, to investigate the alignment paradox between algorithmic optimization and clinical trust. A pyramid-curated dataset enhances model robustness across general, confusing, and human-refined cases. During modeling, SFT provides the foundational reasoning structure, while GRPO and DPO incorporate reward and preference feedback to refine policy behavior, ICL embed two clinical guidance blocks. Model outputs are assessed through a dual-evaluation system combining automatic metrics and doctor-in-the-loop blind assessments. Human experts evaluate outputs on three clinical dimensions—diagnostic accuracy, reasoning clarity, and treatment feasibility—revealing the divergence between benchmark performance and clinician-perceived interpretability in assisted reproductive decision-making.
The proposed dual-alignment framework integrates SFT with three post-training alignment strategies (DPO, GRPO, and ICL) to examine how algorithmic optimization interacts with clinical trust. A pyramid-curated dataset covering general, ambiguous, and expert-refined cases is used to enhance robustness across difficulty levels. SFT establishes the core reasoning structure, while GRPO and DPO introduce reward- and preference-based policy refinement, and ICL incorporates structured clinical guidance blocks. Model outputs are evaluated through a dual-layer assessment combining automatic metrics with triple-blind doctor-in-the-loop reviews. Clinicians rate each output on diagnostic accuracy, reasoning clarity, and treatment feasibility, enabling us to identify divergences between benchmark performance and clinician-perceived interpretability in assisted reproductive decision-making.
\end{flushleft}
\end{figure}
\section{Results}\label{sec2}

\subsection{ Automatic Metrics Performance  
}\label{sec3}
Table 1 summarizes the field-level performance of the four models. Among them, GRPO (the SFT model trained by GRPO) achieves the best overall results in the fields of Infertility type, ART strategy, and COS regimen, with the highest average accuracy (77.14\%) and macro F1 score (50.64\%). Compared to the base SFT model, GRPO improves ART strategy selection accuracy  by 2.68\% and 3.81\% in macro F1. GRPO also demonstrates clear enhancement in the unstructured Initial diagnosis task, achieving a 1.95\% higher partial match rate and a 4.24\% increase in exact matches compared to the baseline SFT. These improvements highlight the value of reward-driven reinforcement learning in capturing complex clinical reasoning. DPO also shows moderate gains in ART strategy choice (accuracy +1.22\% , F1 +2.44\% vs. SFT), confirming that preference-based optimization can strengthen decision alignment. However, its instability is reflected in the initial diagnosis task, where both partial and exact matches drop substantially (41.29\% and 4.63\%, nearly half of SFT), suggesting vulnerability to noisy or ambiguous supervision. Although initialized from the same SFT backbone, ICL failed in nearly all metrics, with only marginal improvement in ART strategy (accuracy +0.95\%). This indicates that simple in-context prompting lacks robustness for multi-layer medical decision-making. Notably, For the COS regimen prediction, all models struggle due to the possible inherently ambiguous category boundaries. GRPO records a slight decrease in accuracy (–1.34\% vs. SFT) and F1 (–0.98\%), while DPO and ICL exhibit a larger accuracy drop (–2.80\%, -4.26\%) despite a minor F1 improvement by DPO (+0.56\%). This suggests that for tasks with complex or flexible category boundaries, even reinforcement learning fails to yield substantial performance gains.
\begin{table*}[t]
\centering
\caption{Field-level performance comparison of SFT, ICL, DPO, and GRPO models on infertility-related decision tasks.}
\label{tab:performence}
\setlength{\tabcolsep}{3pt} % 缩小列距
\renewcommand{\arraystretch}{1.5} % 轻微增行距
\resizebox{\textwidth}{!}{ % 自适应页面宽度
\begin{tabular}{lcccccccccccc}
\toprule
\textbf{Model} &
\multicolumn{2}{c}{\textbf{Infertility Type}} &
\multicolumn{2}{c}{\textbf{ART Strategy}} &
\multicolumn{2}{c}{\textbf{COS Regimen}} &
\multicolumn{2}{c}{\textbf{Average}} &
\multicolumn{2}{c}{\textbf{Initial Diagnosis}} &
\textbf{Gn Dose} \\
\cmidrule(lr){2-3}\cmidrule(lr){4-5}\cmidrule(lr){6-7}\cmidrule(lr){8-9}\cmidrule(lr){10-11}\cmidrule(lr){12-12}

 & ACC & F1 & ACC & F1 & ACC & F1 & ACC & F1 & Partial & Exact & MAE \\
\midrule
SFT  & 92.57 & \textbf{92.18} & 73.81 & 49.18 & \textbf{63.70} & 7.75 & 76.69 & 49.70 & 87.45 & 16.08 & \textbf{43.88 }\\
ICL  & 90.83 & 89.88 & 74.76 & 33.31 & 59.44 & 5.89 & 75.01 & 43.03 & 87.21 & 16.09 & 48.38 \\
DPO  & 92.20 & 91.33 & 75.03 & 46.62 & 60.90 &\textbf{ 8.31} & 76.04 & 48.75 & 41.29 & 4.63  & 45.12 \\
GRPO & \textbf{92.57} & 92.05 & \textbf{76.49} & \textbf{52.99} & 62.36 & 6.87 & \textbf{77.14} & \textbf{50.64} & \textbf{89.40} &\textbf{ 20.32} & 44.94 \\
\bottomrule
\end{tabular}}
\begin{flushleft}
Results are reported for five evaluation fields: infertility type classification, ART strategy selection, COS regimen prediction, initial diagnosis (partial and exact match), and gonadotropin (Gn) starting dose (MAE). ACC = accuracy, F1 = macro F1 score, MAE = mean absolute error. Bold values indicate the best performance in each column. GRPO demonstrates the highest overall accuracy and F1, while DPO shows instability in initial diagnosis.
\end{flushleft}
\footnotesize
\vspace{6pt}
\end{table*}
\subsection{Doctor-in-the-loop Evaluation }
To evaluate the models' clinical performance, we utilized human feedback from an expert panel. For this evaluation, physicians blindly reviewed 100 representative cases, which have the same ART distributions as the test set in automatic evaluation. Here we selected the baseline model (SFT) and the post-training model (GRPO), which achieved the highest automatic scores. Each model's output was scored on four primary dimensions: Diagnosis Accuracy (Acc.), Clinical Reasoning Capability (Reasoning), Treatment Plan Clinical Feasibility (Feasibility), and Hallucination . The results of this expert scoring (Figure 2) revealed a significant discrepancy with the findings from the automated metrics. Despite the automated evaluation showed GRPO consistently outperforming SFT, the physician ratings reversed this trend. Physicians preferred more on the SFT model significantly on both reasoning capability (p = 0.033, adj. p = 0.048, Cohen's d = 0.22) and clinical feasibility (p = 0.015, adj. p = 0.045, Cohen's d = 0.25). There were No significant differences observed in diagnosis accuracy between the two models. For the additional dimension, hallucination, GRPO was rated to have a lower rate (15.00\%) compared to SFT (18.61\%). Though GRPO’s lower hallucination rate and superior automated metrics, the findings on reasoning and feasibility suggest that the algorithmic improvements did not translate into perceived clinical trust or practical usefulness. Together, these results clearly demonstrate that algorithmic gains from reinforcement-style alignment do not translate into improvements in clinical trust. Instead, they reveal an alignment paradox in which optimization enhances benchmark performance but degrades reasoning clarity and perceived decision feasibility.
\begin{figure}[t]
\centering
\captionsetup{justification=raggedright,singlelinecheck=false}
\renewcommand{\arraystretch}{1.05}
\setlength{\tabcolsep}{2.2pt}
\caption{\textbf{Doctor-in-the-loop evaluation.}}

\begin{minipage}[t]{0.38\textwidth}
\centering
\textbf{(a)} Response subjective evaluation\\[2mm]
\begin{tabular}{lcccc}
\toprule
Model & Acc. & Reason. & Feas. & Hallu. (\%) \\
\midrule
SFT  & \textbf{4.04} & \textbf{4.17} & \textbf{4.21} & 18.6 \\
GRPO & 3.99 & 4.08 & 4.09 & \textbf{15.0} \\
\bottomrule
\end{tabular}
\end{minipage}
\hfill
% 右图：间距修正 + 压缩 + 美化
\begin{minipage}[t]{0.57\textwidth}
\centering
\textbf{(b)} Winning rate comparison.\\[2mm]
\begin{tikzpicture}[every node/.style={scale=0.9}]
\begin{axis}[
    ybar,
    bar width=14pt,          
    width=0.9\textwidth,
    height=3.7cm,             
    ymin=0, ymax=60,
    symbolic x coords={G.T., GRPO, SFT},
    xtick=data,
    ylabel={Winning Rate (\%)},
    tick label style={font=\footnotesize},
    label style={font=\footnotesize},
    nodes near coords,
    every node near coord/.append style={font=\footnotesize, anchor=south},
    enlarge x limits=0.12,     % 让GT不贴y轴
    ymajorgrids=true,
    grid style=dashed,
    axis line style={thick},
    tick align=inside,
]
\addplot[fill=blue!55!gray] coordinates {(G.T.,22.7) (GRPO,26.2) (SFT,51.2)};
\end{axis}
\end{tikzpicture}
\end{minipage}

\begin{flushleft}
Results of (a) expert-rated performance and (b) blind winning rate comparison across SFT, GRPO, and physician ground truth (G.T.). (a) The SFT shows significantly higher reasoning and feasibility scores, GRPO has a lower hallucination percentage. (b) both SFT and GRPO outperform the human-level answer, SFT accounts for more than half of the better responses.
\end{flushleft}

\label{fig:doctor_combined}
\end{figure}

\subsection{Winning rate}
In addition to the clinical dimensions, the expert panel was asked to blindly identify the overall best response from the scored two responses and a third human-level response, which we directly use the physician ground truth. The results were striking. Physicians selected the SFT model in more than half of the cases (51.2\%), far exceeding both the GRPO answers (26.2\%) and even the original treatment plans (22.7\%). Interestingly, even the GRPO, despite being less favored than SFT, still surpassed the ground truth. This outdoing suggests that both aligned models may generate decisions perceived as comparably or even more clinically useful than those originally made by human physicians. This pattern enhances the claim of the divergence between algorithmic optimization and clinical perception. While GRPO achieved promising token-level accuracy and lower hallucination rates, physicians overwhelmingly preferred the SFT model for its reasoning transparency and decision coherence. In other words, though improves benchmark metrics, the optimization appeared to erode human-perceived comprehension and trust.

Collectively, these results suggest that current alignment objectives may not fully capture what clinicians value in practice, such as clarity, rationale, and contextual adaptability. Both SFT and GRPO have already exceeded human consistency, yet differ sharply in how they earn clinical trust. This pattern demonstrates that alignment optimization may shift the model toward high-scoring but less interpretable solutions, widening the gap between algorithmic metrics and human-centered evaluation

\subsection{Subgroup Analysis}
\begin{figure}
    \centering

    \includegraphics[width=1\linewidth]{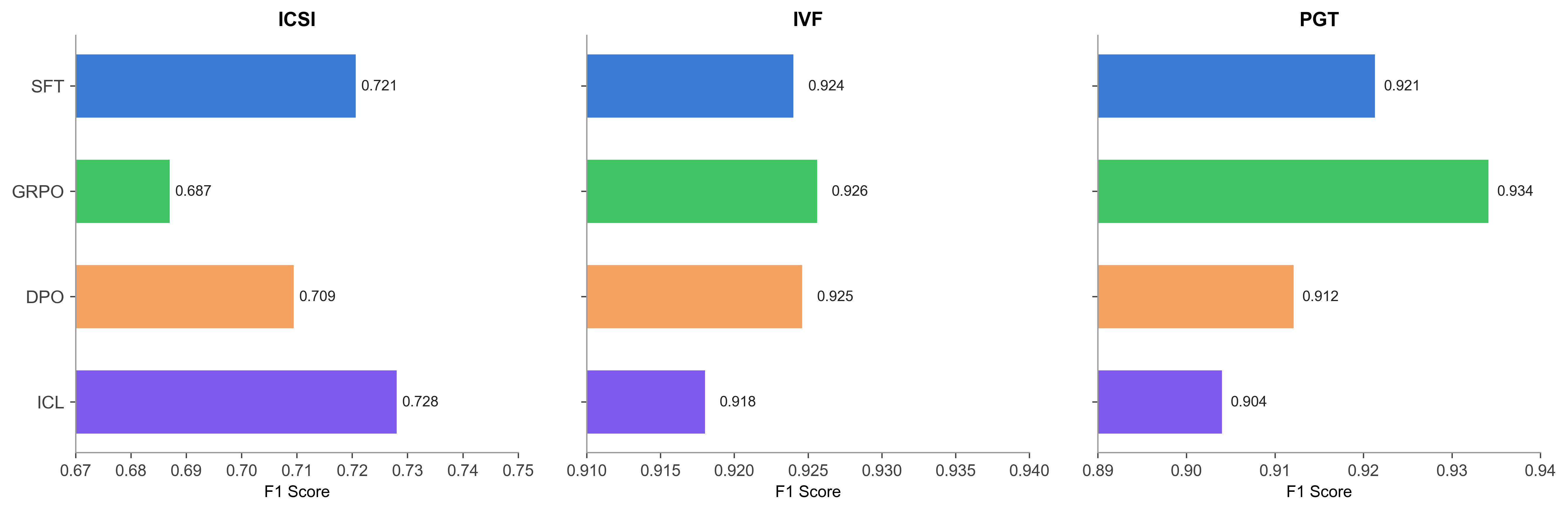}
    \caption{Reinforcement learning–based alignment improves F1 scores in IVF and PGT, but consistently reduces performance in ICSI. All values are reported as percentages.}
    \label{fig:3}
\end{figure}

To elucidate how different alignment strategies influence decision performance across heterogeneous clinical contexts, we first evaluated model behavior at the level of three major ART categories: ICSI, IVF, and PGT (Figure \ref{fig:3}). Both GRPO and DPO yield marginal yet consistent gains over SFT in IVF (SFT = 0.924, GRPO = 0.926, DPO = 0.925), while GRPO further enhances PGT performance (SFT = 0.921 → 0.934). In contrast, reinforcement learning–based optimization results in systematic degradation for ICSI (0.721 → 0.687 for GRPO; 0.709 for DPO), suggesting that reward-driven alignment may distort the decision boundaries underlying this highly constrained treatment category. Prompt-based guidance (ICL) offers only slight improvement in ICSI (0.721 → 0.728), indicating that contextual augmentation alone is insufficient to resolve the intrinsic reasoning difficulty of this subgroup.

\begin{figure}
    \centering
    \caption{Differences between GRPO and SFT in Subtype Analysis}
    \includegraphics[width=1\linewidth]{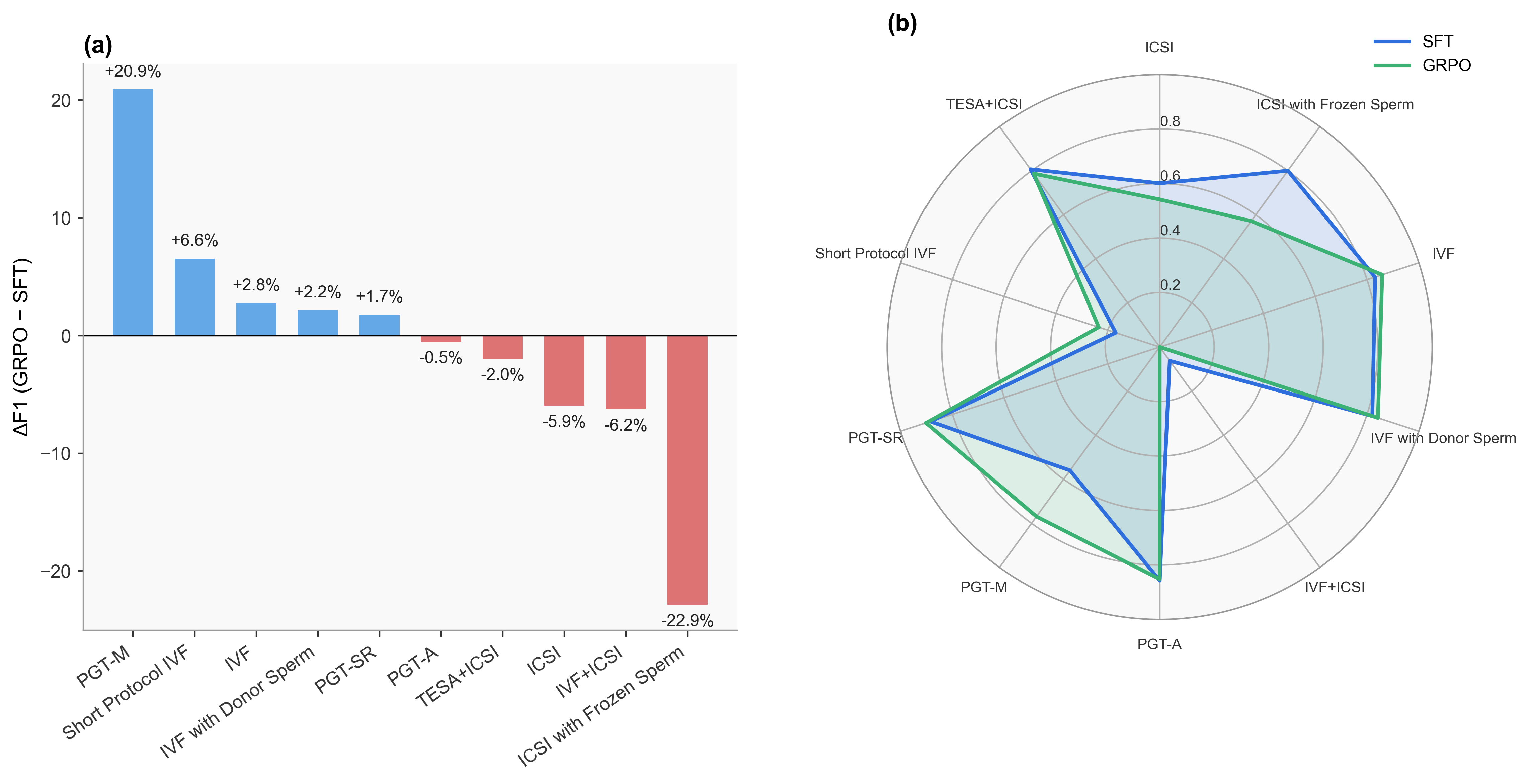}
\begin{flushleft}
a. Fine-grained ART subtype performance comparison between SFT and GRPO. Each bar shows the change in F1 score ($\Delta$ F1 = GRPO − SFT, \%) across eleven ART subtypes, covering ICSI, IVF, and PGT variants. GRPO exhibits marked declines across all ICSI-related categories but delivers substantial gains in several complex or long-tail subgroups, most notably PGT-M (+20.9\%) and short-protocol IVF (+6.55\%). b. ART subtype performance comparison of SFT and GRPO. The ICSI-with-Donor-Sperm subtype is excluded due to its single-sample size. 
\end{flushleft}

    \label{fig:4}
\end{figure}

To probe the structural origins of these trends, we conducted a fine-grained analysis across eleven ART subtypes using SFT and GRPO (Table \ref{tab:subtype}, Figure \ref{fig:4}). The detailed subtype results mirror the aggregated patterns. GRPO underperforms across all ICSI-related subgroups, with the most substantial decline observed in the dominant standard ICSI subtype (−5.93\%). Meanwhile, GRPO produces marked gains in clinically complex or long-tail PGT subtypes, most notably PGT-M (+20.9\%) and to a lesser extent PGT-SR (+1.74\%), demonstrating that reinforcement-style optimization can reduce error modes associated with sparse or heterogeneous data distributions. A similar pattern emerges for IVF subtypes: GRPO produces remarkable improvements in short-protocol IVF (+6.55\%) and a moderate gain in standard IVF (+2.75\%). The apparent decline in ICSI with frozen sperm should be interpreted cautiously due to extremely limited sample size (n = 3).

Collectively, these findings indicate a distinct trade-off introduced by RL-based alignment. While GRPO enhances robustness in underrepresented or clinically intricate subgroups, it simultaneously compromises stability in well-represented, tightly structured categories such as ICSI. This divergence underscores a central challenge for clinical LLM alignment: optimizing for long-tail reasoning without distorting high-confidence regions of the decision space. Designing clinically reliable alignment strategies therefore requires balancing global performance with local subgroup fidelity—an essential consideration for decision-critical medical applications.

\begin{table}[t]
\centering
\caption{\textbf{Fine-grained ART subtype performance.} }
\label{tab:subtype}
\setlength{\tabcolsep}{6pt}
\renewcommand{\arraystretch}{1.2}
\begin{tabular}{lcccccc}
\toprule
\textbf{ART Subtype} & \multicolumn{3}{c}{\textbf{SFT}} & \multicolumn{3}{c}{\textbf{GRPO}} \\
\cmidrule(lr){2-4} \cmidrule(lr){5-7}
 & \textbf{Precision (\%)} & \textbf{Recall (\%)} & \textbf{F1 (\%)} & 
   \textbf{Precision (\%)} & \textbf{Recall (\%)} & \textbf{F1 (\%)} \\
\midrule
ICSI         & 57.01 & 63.54 & \textbf{60.10}& 54.17 & 54.17 & 54.17 \\
ICSI (DS)    & 0.00  & 0.00  & 0.00  & 0.00  & 0.00  & 0.00  \\
ICSI (FS)    & 100.00 & 66.67 & \textbf{80.00}& 50.00& 66.67& 57.14 \\
IVF          & 79.92 & 86.54 & 83.10 & 80.69& 91.72& \textbf{85.85}\\
IVF (DS)     & 88.89 & 76.19 & 82.05 & 94.12& 76.19& \textbf{84.21}\\
IVF+ICSI     & 6.25  & 6.25  & \textbf{6.25}& 0.00  & 0.00  & 0.00  \\
PGT-A        & 95.45 & 77.78 & \textbf{85.71}& 88.46& 82.14& 85.19 \\
PGT-M        & 77.78 & 43.75 & 56.00 & 1.00& 62.50& \textbf{76.92}\\
PGT-SR       & 83.64& 93.88& 88.46& 85.19& 95.83& \textbf{90.20}\\
IVF (short)  & 30.00& 11.84& 16.98& 46.15& 15.79& \textbf{23.53}\\
TESA+ICSI    & 86.21& 75.76& \textbf{80.65}& 85.71& 72.73& 78.69\\
\bottomrule
\end{tabular}
\begin{flushleft}
 All values are reported as percentages. This table compares precision, recall, and F1 scores between the SFT and GRPO models across eleven ART subtypes, with higher F1 values highlighted in bold. The results illustrate a trade-off between overall stability and targeted optimization, suggesting that reinforcement-style alignment may improve performance in select subgroups while reducing consistency in others.
\end{flushleft}
\end{table}

\section{Methods}\label{sec11}

In this section, we will first define the task dealt by the LLMs, then detail the dataset curation and model training process. Finally, we will handle the evaluation metrics. As we mentioned in the introduction, we are expected to explore three main questions: the performance of SFT baseline and various post-training alignment strategies, the contribution pattern of the alignment, and the tension between algorithm alignment and clinical trust.

\subsection{Data Source \& Ethic Considerations}

We collected the electronic healthcarae records (EHRs) from West China Second University Hospital spanning from January 2020 to December 2022, consisting of 19800 patients’ information. The study was approved by the Ethics Committee of West China Second University Hospital, Sichuan University (ID: 2022288). The EHRs have been manually reviewed and corrected to ensure data accuracy. All EHR data used in this study were fully de-identified before being accessed by the research team. The de-identification procedure followed HIPAA Safe Harbor standards, including the removal of all direct identifiers (e.g., name, date of birth, medical record number, contact information, provider information) and all quasi-identifiers (e.g., dates, locations, and institutional identifiers). Only aggregated clinical descriptors necessary for the reasoning task (e.g., high-level patient history, laboratory summaries) were retained. Access to the de-identified dataset was restricted to authorized study personnel through institution-managed credentials and encrypted storage. Reviewers performing the blinded evaluation accessed only the de-identified clinical vignettes and model-generated reasoning content through a secure, read-only interface; no downloads or re-identification attempts were permitted. All access was logged and monitored by an internal auditor to ensure compliance with institutional clinical data governance policies.

After manually proofreading and data cleaning, a final corpus of 8201 electronic healthcare records was utilized as our basic dataset with patient mean age = 31.79, s.d. = 4.63. All the confidential information, including patient name and code, was masked to protect the privacy. Grouped by ART, there are in total 11 different methods, which can be generally assigned to three ART generations: In Vitro Fertilization (IVF), Intracytoplasmic Sperm Injection (ICSI), and Preimplantation Genetic Testing (PGT). Since the WHO has regulated the name of PGT and its subtypes, we convert all the outdated names, such as PGD (now PGT-M), PGS (now PGT-A), to the stipulated one. Statistically, there are 71.21\% IVF (n = 5840), including 59.6\% standard IVF (n = 4885), 8.17\% Short Protocol IVF (\textit{Short-time insemination}, n = 670), 3.48\% IVF with Donor Sperm (n = 285); 17.92\% ICSI (n = 1470), including 11.57\% standard ICSI (n = 949), 3.68\% TESA+ICSI (n = 302), 1.8\% IVF+ICSI (n = 155), 0.65\% ICSI with Frozen Sperm (n = 53), and 0.13\% ICSI with Donor Sperm (n = 11); 10.86\% PGT (n = 891), including 5.12\% PGT-SR (n = 420), 3.76\% PGT-A (n = 308), and 2.12\% PGT-M (n = 163).  For COS regimen, there are 12 categories: GnRH Antagonist Fixed Protocol (Antagonist-Flex), Luteal Short-Acting Long Protocol (Luteal-Short), GnRH Antagonist Flexible Protocol (Antagonist-Fixed), Follicular Phase Long-Acting Protocol (Long-Acting), Progestin-Primed Ovarian Stimulation Protoco (PPOS), Clomiphene Citrate Plus Gonadotropins (CC+Gn), Mild Stimulation Protocol / Direct Gn Protocol (Mild/Direct Gn), Luteal Phase Stimulation Protocol (Luteal-Stim), Clomiphene or Letrozole Combined with Gonadotropins (CC/Letro+Gn), Conventional Ultra-Long Protoco (Ultra-Long), Modified Ultra-Long Protocol (Mod-ULong), Short Protocol (GnRH-a Short Protocol) (Short).

\subsection{Dataset Curation}
\subsubsection{SFT Dataset Construction  }
The SFT dataset requires training pairs of prompt inputs and ground truth outputs. For the input prompt, we collated nine fields per patient, ranging from structured baseline data to unstructured textual descriptions. The structured data included Female Age, Menstrual Cycle, Weight, BMI, AMH, FSH, and Infertility Years. The unstructured annotations included gynecology ultrasound reports and medical history, which typically documents prior conditions such as surgical history, assisted reproduction history, and male semen analysis. Since all annotations were in Chinese, we performed professional translation using ChatGPT-4o. For the ground truth output, we designed a double-layer structure to ensure explainability: (1) a multi-part clinical reasoning (Chain-of-Thought, CoT), and (2) the final diagnosis and treatment plan.
However, Manually annotating thousands of clinical reasoning chains was neither time-permitted nor affordable. Therefore, we generated the CoT component using an in-context learning (ICL) approach with a diverse case boutique prompt as Few-shot prompt, a method we previously validated for clinical reliability\cite{liu_reliability_2025}. This boutique set included six commonplace ART cases, with sample CoTs carefully curated by two expert-level physicians (these cases were excluded from our basic dataset). The resulting CoTs were structured into four aspects: Diagnosis reasoning, Assisted reproduction technology decision, Ovarian stimulation protocol selection, and Gn initiation dosing rationale. These four are appropriately apposite to part 2, diagnosis \& treatment plan. In the second part, we set five final answer fields to mimic the physician’s final decision-making annotation: 
 \begin{itemize}
     \item \textbf{Diagnosis Fields:} Infertility Type (Primary, Secondary, or Other) and Initial differential diagnosis .
     \item \textbf{Treatment Fields:} ART strategy (11 subtypes), COS regimen (12 different protocols), and the Gn starting dose,

For diagnosis, we have Infertility Type judgment and Initial differential diagnosis. Each case was labeled with one of three infertility categories: Primary Infertility, Secondary Infertility, or Other (unclear or multifactorial cases). Initial diagnosis mainly focuses on the female side’s potential causes, and the male side is included if applicable. For the treatment plan, we pose the ART strategy, the COS regimen, and the Gn starting dose. These three vital plans build the foundation of the following assisted reproduction and can be inferred through the comprehensive input information. The COS regimen also has 11 different protocols. In our training process, we separate the entire dataset into the train set (80\%), the validation set (10\%), and the test set (10\%)
 \end{itemize}

\subsubsection{Pyramid Dataset Construction}

To further strengthen post-training alignment and improve the model’s discrimination on clinically ambiguous or low-frequency cases, we constructed a \textit{pyramid-style} alignment dataset. This dataset specifically addresses two key limitations of the SFT baseline. Specifically, its bias toward dominant treatment categories and its reduced robustness in rare or clinically complex scenarios. The pyramid consists of three layers, each designed to provide distinct alignment signals: \textbf{general enhancement}, \textbf{confusion enhancement}, and \textbf{human enhancement}. After SFT, we performed model inference on both training and test sets to obtain paired ground-truth and model-generated responses. For DPO, which relies on pairwise preference supervision, ground-truth outputs served as preferred responses, whereas SFT-generated outputs were used as rejected responses. GRPO utilized the same dataset distribution but operated directly on prompt–response pairs. The full alignment dataset was then assembled in a top-down manner as follows:

\textbf{Human Enhancement} 

  This layer focuses on the most clinically challenging and sparsely represented cases. Two representative long-tail categories were selected: (1) \textit{IVF + ICSI}, a hybrid of two ART generations that accounts for only 1.8\% of SFT training data; and (2) \textit{Short-protocol IVF},a complicated variant of standard IVF that is difficult even for specialists. A total of Fifty cases (25 per category) were curated to maximize signal clarity and expose the model to high-value reasoning patterns rarely encountered in routine data.

\begin{figure}
    \centering
    \caption{Cross-metric confusion analysis for COS regimen and ART prediction.}
    \includegraphics[width=1\linewidth]{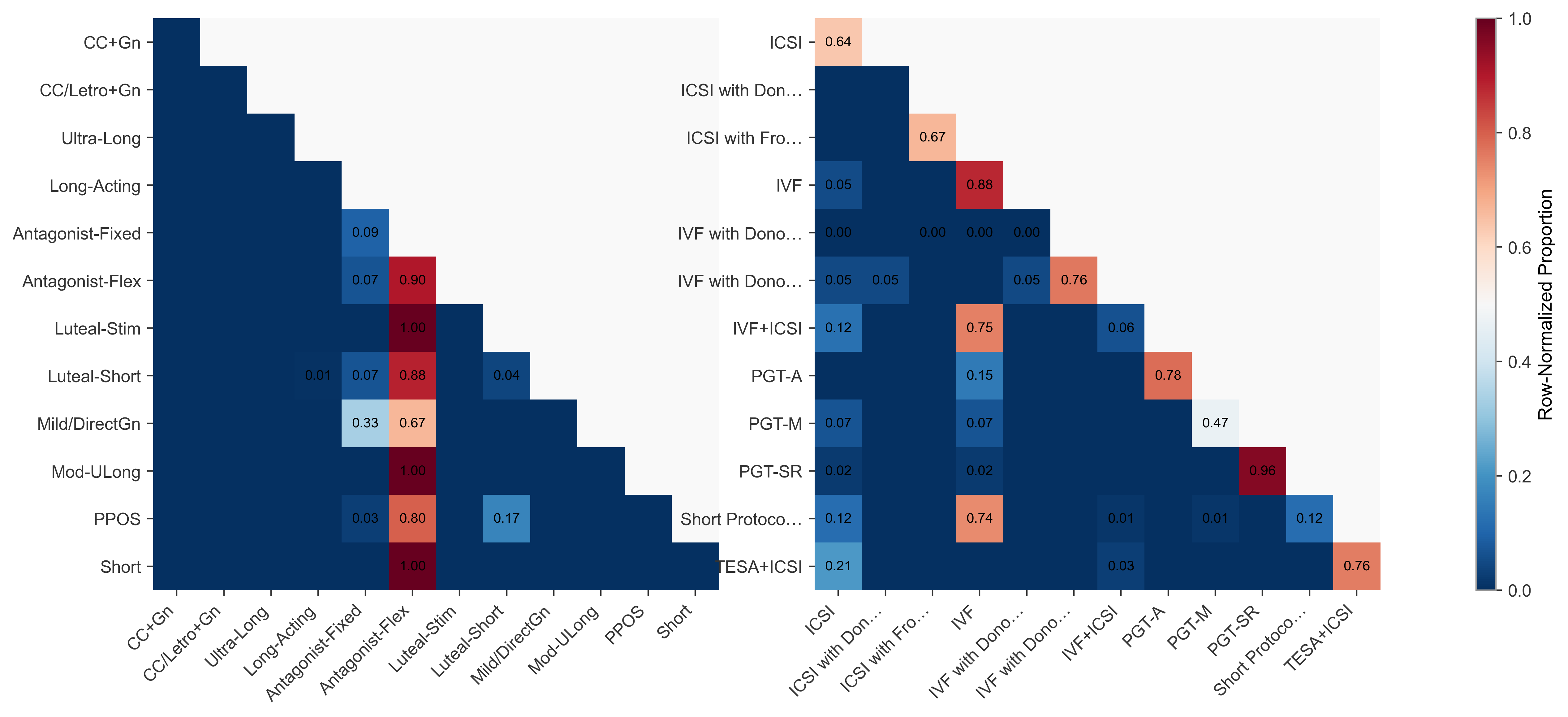}
\begin{flushleft}
(a) Row-normalized lower-triangle confusion matrix showing the correspondence between ground-truth and model-predicted controlled ovarian stimulation (COS) regimens. (b) Row-normalized lower-triangle confusion matrix for treatment type predictions (IVF, ICSI, and PGT). Each cell represents the proportion of samples predicted as a given category among all samples with the same ground-truth label, with color intensity indicating normalized frequency (blue = under-prediction, red = over-prediction). These matrices reveal systematic confusion patterns, particularly the model’s tendency to over-predict dominant categories such as the Antagonist regimen and IVF. These patterns directly informed the construction of the \textit{pyramid} alignment dataset, where mid-layer samples were drawn from high-confusion regions to enhance discriminative robustness.
\end{flushleft}

    \label{fig:5}
\end{figure}

\textbf{Middle layer: Confusion Enhancement} 
This layer focuses on systematic confusion observed in the predictions of the initial SFT model. We evaluated the outputs across both ART strategy and COS regimen, using row-normalized confusion matrices (Fig\ref{fig:5}) to identify high-confusion regions. A total of 628 samples were extracted from categories with individual error rates greater than 10\%. Particularly, poor discrimination among non-Antagonist COS regimens led to the inclusion of all other regimen types, allowing this layer to fully capture the blind spots of the model.

\textbf{Bottom layer: General Enhancement} 
To avoid overfitting toward rare or confusing samples, the base layer includes a balanced mixture that includes all groups of cases. This bottom layer contains 700 samples with at least one incorrect field and 300 fully correct samples across the ART strategy and COS regimen tasks, providing stable generalization support and preventing distributional skew introduced by the upper layers.

In total, the alignment dataset consists of 1,678 samples distributed across the three layers, and is divided into 80\%, 10\%, and 10\% for training, validation, and testing, respectively. Although the top-layer samples contributes a small proportion of the dataset, their high informational density offset the dilution from lower layers. This allows the model to internalize complex reasoning and rare clinical patterns effectively. For GRPO training, we adopted the same alignment dataset to ensure a consistent task distribution. However, since GRPO does not rely on pairwise preference supervision, each pair was processed by removing the rejected response and retaining only the ground-truth response. The same 80/10/10 train–validation–test split was applied. 

\subsection{Modeling}

To investigate how different alignment paradigms influence clinical reasoning and decision-making, we compare representative strategies built on a shared backbone. Given the clinical nature of our task, we opted not to directly fine-tune a general-purpose pre-trained model such as Qwen-2.5 or LLaMA-3. Instead, we utilized OpenBioLLM-8b\cite{noauthor_aadityallama3-openbiollm-8b_nodate}, a domain-specific open-sourced model tailored for the medical field, which has demonstrated strong performance compared to other state-of-the-art open-sourced models at the time of use. OpenBioLLM is built upon LLaMA-3 and has been trained on a range of medical knowledge sources. It has been adopted in several studies across clinical NLP and vision-language applications\cite{shi_fine-tuning_2025,alshibli_vision-biollm_2025}.  (OpenBioLLM-8B). We evaluate four alignment paradigms: SFT, DPO, GRPO, and ICL, chosen to represent supervised, preference-based, reinforcement-based, and prompt-based alignment strategies. These four paradigms jointly span the major alignment families used in contemporary LLM research, enabling us to systematically assess how algorithmic optimization interacts with clinical reasoning and trust.

In this study, we aim to obtain a LLM with specific capability in infertility diagnosis and treatment planning, accompanied by the reasoning text, or Chain-of-Thought (CoT) on each aspect. We define infertility diagnosis and treatment planning as a structured, multi-output reasoning problem. Given a patient record \( X = \{x_1, x_2, \dots, x_n\} \), which integrates both structured baseline data and unstructured clinical narratives (e.g., gynecology ultrasound reports and medical history), the model \( f_\theta \) is required to generate five interdependent clinical decisions and their corresponding reasoning chains. Formally, the task can be expressed as learning a mapping function:
\[
f_\theta: X \rightarrow (Y_1, Y_2, Y_3, Y_4, Y_5, C),
\]
where \( (Y_1, Y_2, Y_3, Y_4, Y_5) \) represent five structured outputs:

(1) infertility type judgment, (2) initial diagnosis, (3) assisted reproductive technology (ART) strategy, (4) controlled ovarian stimulation (COS) regimen, (5) gonadotropin (Gn) starting dose.

\( C \) denotes the chain-of-thought (CoT) reasoning text that explicitly supports and explains each decision. The learning objective of the model is to maximize the joint likelihood of generating both the 
reasoning process and the final structured decisions conditioned on the patient data:
\[
\mathcal{L}(\theta) = \sum_{(X, Y_1,\dots,Y_5,C)\in\mathcal{D}} 
\log P_\theta(Y_1,\dots,Y_5,C \mid X).
\]
In practice, this formulation provides a unified objective for multiple alignment paradigms, including SFT, DPO, and GRPO. It ensures that the model learns not only to produce correct clinical outcomes,  but also to generate transparent and clinically faithful reasoning chains. Based on this objective function, we trained the models and evaluated their performance through a dual evaluation pipeline that incorporated both algorithmic metrics and human expert feedback. Figure \ref{fig:Overview} illustrates the whole process.  

\textbf{SFT}

This stage serves as the foundation for all subsequent alignment strategies, providing a clinically coherent policy network pretrained with structured reasoning supervision. In the Supervised Fine-Tuning (SFT) stage, we train the base model on the entire training set, where each instance consists of a structured clinical description and a corresponding expert-annotated response including reasoning and diagnosis, as laid out in data section. Training was performed with LoRA (learning rate = 3e-5, batch size = 4) for 10 epochs on a single A100 GPU. The resulting model serves as the reference policy for the subsequent DPO, GRPO, and ICL stages.

\textbf{DPO }

This variant focuses on aligning model reasoning with expert preferences by directly optimizing token-level log-likelihood contrast between preferred and dispreferred responses. Direct Preference Optimization (DPO) is a critic-free reinforcement learning method that has recently emerged from the RLHF (Reinforcement Learning from Human Feedback) paradigm\cite{rafailov_direct_2023} It provides a relatively simple and stable alternative to reward-model-based offline approaches for aligning language models with human preferences and has shown promising results in reducing undesirable behaviors in baseline models. Unlike traditional RLHF methods that rely on learning a separate reward model or value function, DPO directly optimizes the policy by contrasting preferred and dispreferred responses. Specifically, each training data point consists of a prompt    \textit{x}, a preferred response $y_{w}$ , and a less-preferred response $y_{l}$  \textit{.} Given the prompt \textit{x}  , the DPO loss encourages the policy model to assign a higher likelihood to $y_{w}$ over $y_{l}$\textit{,} thereby aligning the model outputs more closely with human preferences. The DPO loss function is formulated as:

\begin{equation}
\mathcal{L}_{\mathrm{DPO}}(\pi_{\theta}; \pi_{\mathrm{ref}}) 
= - \mathbb{E}_{(x, y_w, y_l) \sim \mathcal{D}}
\left[
    \log \sigma \left(
        \beta \left[
            \log \frac{\pi_{\theta}(y_w \mid x)}{\pi_{\mathrm{ref}}(y_w \mid x)} 
            - 
            \log \frac{\pi_{\theta}(y_l \mid x)}{\pi_{\mathrm{ref}}(y_l \mid x)}
        \right]
    \right)
\right]
\end{equation}

Where $\pi_{\theta}$ is the current policy model, ${\sigma}$  denotes the sigmoid function, and ${\beta}$  is a temperature parameter controlling the sharpness of the preference. Notably, this loss assigns equal weight to all tokens in each response, which is precisely what we aim to achieve—aligning the entire reasoning process and final diagnosis to the human level. Since we have access to the correct final answer, this setup encourages the model not only to predict the correct outcome but also to generate a coherent and clinically sound chain of thought leading to it. In the context of our study, where both the intermediate reasoning (e.g., identifying relevant clinical clues) and the final decision (e.g., treatment plan) are critical, such token-level uniform supervision helps ensure that the model learns to reflect expert-like logic across the entire response, rather than merely optimizing for the final output token. In our experiments, we use the SFT model described in Section 2.3 as the reference policy. DPO was also conducted using an A100 with beta = 0.3, learning rate = 3e-7, batch size = 8 for 1 epoch. 

\textbf{GRPO} 

This variant leverages reinforcement-based relative optimization to enhance decision consistency while eliminating the computational overhead of value-function training. Group Relative Policy Optimization (GRPO) is an evolutionary variant of Proximal Policy Optimization (PPO)\cite{schulman_proximal_2017}. For PPO, its advantage is computed by applying Generalized Advantage Estimate (GAE)\cite{schulman_high-dimensional_2018}, based on the rewards and a learned value function. Consequently, a value function needs to be trained alongside the policy model and using a per-token KL penalty from the reference model to mitigate over-optimization of the reward model. As the value function is involved in the training, it is typically another model of comparable size to the policy model, bringing a gargantuan computational burden. While for GRPO, it obviates an additional value function for advantage computation and instead uses the reward average of multiple outputs sampled from the same question as the baseline. More specifically, for each question ${q}$, GRPO samples a group of outputs $\{o_1, o_2, \cdots, o_G\}$ from the old policy $\pi_{\theta}$ and then optimizes the policy model by maximizing the following objective:
\begin{equation}
\begin{aligned}
J_{\mathrm{GRPO}}(\theta) 
&= \mathbb{E}_{q \sim P(Q), \{o_i\}_{i=1}^{G} \sim \pi_{\theta_{\mathrm{old}}}(O|q)} 
\Bigg[
\frac{1}{G} \sum_{i=1}^{G} 
\frac{1}{|o_i|} 
\sum_{t=1}^{|o_i|} 
\Big\{
\min \Big[
\frac{\pi_{\theta}(o_{i,t} \mid q, o_{i,<t})}
{\pi_{\theta_{\mathrm{old}}}(o_{i,t} \mid q, o_{i,<t})} 
\hat{A}_{i,t}, \\
&\quad\quad\quad\quad
\mathrm{clip}\Big(
\frac{\pi_{\theta}(o_{i,t} \mid q, o_{i,<t})}
{\pi_{\theta_{\mathrm{old}}}(o_{i,t} \mid q, o_{i,<t})}, 
1-\epsilon, 1+\epsilon
\Big)\hat{A}_{i,t}
\Big]
- \beta D_{\mathrm{KL}}[\pi_{\theta} \,\|\, \pi_{\mathrm{ref}}]
\Big\}
\Bigg]
\end{aligned}
\end{equation}
where $\epsilon$ and $\beta$ are hyperparameters, and $\hat{A}_{i,t}$ is the advantage calculated based on relative rewards of the outputs inside each group only, which will be detailed in the following subsections. GRPO offers a compelling solution for medical domains where training a value function is often impractical due to limited data, and where structured, interpretable supervision is essential for aligning model behavior with expert expectations.

Reward design follows the common practices\cite{shao_deepseekmath_2024} by using the accuracy reward, aimed at achieving the "algorithm alignment". Since we have four structured final answer fields, except for the initial diagnosis which comprises long sentences, we used a combined final reward to aggregate four separate rewards:

\[
r_i = \lambda_{\mathrm{IT}} r_i^{\mathrm{IT}} + 
       \lambda_{\mathrm{COS}} r_i^{\mathrm{COS}} +
       \lambda_{\mathrm{Gn}} r_i^{\mathrm{Gn}} +
       \lambda_{\mathrm{ART}} r_i^{\mathrm{ART}}.
\]

The weight for each field was set to:
\[
\lambda_{\mathrm{IT}} = 0.2, \quad
\lambda_{\mathrm{COS}} = 0.3, \quad
\lambda_{\mathrm{Gn}} = 0.2, \quad
\lambda_{\mathrm{ART}} = 0.3.
\]

The weighting coefficients were determined empirically based on the relative clinical importance and information density of each decision field, with higher weights assigned to COS and ART strategies that directly affect treatment outcomes. Additionally, as the Gn dose is a numerical answer, to evaluate the predicted Gn starting dose, we designed a graded reward function that reflects real-world clinical practices. In clinical ovarian stimulation protocols, the Gn dose is typically adjusted in increments of 25 IU, making a deviation of $\leq 25$ IU clinically acceptable. Therefore, we assign:

\[
r_{\mathrm{Gn}} =
\begin{cases}
1.0, & \text{if } | \hat{y} - y | \leq 25,\\[4pt]
0.5, & \text{if } 26 \leq | \hat{y} - y | \leq 50,\\[4pt]
0.0, & \text{otherwise.}
\end{cases}
\]

This tiered reward encourages the model to approximate the clinically preferred range, even if the exact value is not matched, and reflects the tolerance physicians often exhibit during dose adjustment in practice.

\textbf{ICL - Guideline-Based Prompt Alignment}

This variant enhances reasoning generalization through guideline-based prompts. In-context learning (ICL) is a rapid, training-free post-alignment approach commonly applied to general-purpose pretrained models. Unlike conventional few-shot ICL, which relies on instance-level exemplars, our variant repurposes clinical heuristic maps into textual guidance blocks. These guidelines are distilled from physicians’ decision flowcharts, translating structured expert reasoning paths (e.g., stimulation protocol selection, ART strategy prioritization) into natural-language rules embedded within the prompt. This design transforms symbolic medical knowledge into interpretable prompt-based supervision, allowing the model to internalize domain heuristics without additional parameter updates. In this variant, We proposed two guidance on ART strategy and COS regimen. The SFT model is retained as the inference backbone because of its domain-specific alignment, particularly its proficiency in maintaining output format consistency and structured clinical reasoning. For each test case, two such guideline blocks are inserted into the prompt (examples provided in the Supplemental File). Additionally, we prepend the instruction:

\textit{\textbf{“DO NOT USE THE GUIDELINE UNLESS YOU ARE NOT SURE ABOUT YOUR ANSWER.”}}

This design encourages the model to rely primarily on its internalized reasoning ability and consult the guideline only as a fallback when uncertain. In this way, the instructional ICL serves as a lightweight and interpretable alternative to gradient-based alignment, effectively simulating real-world physician behavior—where decisions are primarily experience-driven, with reference to protocols only when ambiguity arises.

\subsection{Dual-Evaluation Protocol}
We employed a dual-layer evaluation framework to assess both quantitative task performance and qualitative clinical quality.  This framework integrates (1) automatic field-level metrics and (2) triple-blind doctor-in-the-loop assessment. This design ensures comprehensive examination of both algorithmic accuracy and clinician-perceived interpretability.

\subsubsection{Automatic Evaluation}

Model outputs were evaluated using standard metrics for structured clinical fields. Specifically, we compute Accuracy and Macro-F1 for categorical predictions, including Infertility Type, ART Strategy, and COS Regimen. Because the initial diagnosis field exhibits high variability and paraphrasing, exact word-by-word matching is infeasible and unreliable. Therefore, an auxiliary large language model was used as a semantic judge to determine whether the generated diagnosis semantically entailed the ground-truth diagnosis. We report the precision of containing at least one true diagnosis or all of the ground truth. For the numerical field of Gn starting dose, prediction error was quantified using Mean Absolute Error (MAE).  Together, these metrics assess the model’s structured decision-making performance but do not capture the quality of its reasoning process.

\subsubsection{Doctor-in-the-loop Evaluation}

To evaluate clinical reasoning and decision-making quality beyond quantitative metrics, we conducted a domain-expert assessment. This evaluation addressed two core questions: 

(1) Does post-training alignment improve reasoning coherence and clinical plausibility?

(2) Can aligned models approximate expert-level performance?

Experts rated each output along four clinically grounded dimensions:

\begin{itemize}
    \item Clinical Reasoning Capability: whether the model follows a medically sound thought process;
    \item Diagnosis Accuracy: whether the inferred Infertility type and  Initial diagnostic are correct or contain false positive predictions;
    \item Treatment Feasibility:  whether the entire recommended treatment plan is medically appropriate and consistent with clinical practice;
    \item Hallucination: whether the reasoning process contains irrelative or fabricated information.
\end{itemize}

These dimensions cover all the blocks in the response and can be seen as a comprehensive subjective evaluation. Each output is scored on a 5-point Likert scale (1 = poor, 5 = excellent) according to the rubrics defined above. For each evaluation case, we compare the outputs of the SFT model and the best-performing post-alignment model. 
Additionally, to further benchmark model performance against clinical standards, we include a third response generated from the ground truth answer, representing the expert-level response. The graders are then asked to select the overall best response among the three. All responses are blindly evaluated: the graders are unaware of the matching relationship, thereby reducing bias. Each case is independently reviewed by a panel of board-certified physicians with experience in reproductive medicine. Considering the limited time available for physicians, we selected 100 representative cases as the evaluation set, whose category distribution was kept consistent with that of the test set. 

\subsection{Statistical information}

For automatic evaluation, we reported Accuracy, macro-F1, and Mean Absolute Error (MAE) across all structured decision fields. Accuracy and macro-F1 were computed for categorical outputs (e.g., infertility type, ART strategy, and COS regimen), while MAE was used to quantify the deviation of numerical predictions (e.g., gonadotropin starting dose) from the ground truth. These metrics were calculated on the held-out test set and compared across alignment strategies (SFT, DPO, GRPO, and ICL) without inferential testing, as they were derived from deterministic model outputs. For physician evaluations, ratings on three clinical dimensions: Diagnosis Accuracy, Clinical Reasoning Capability, and Treatment Plan Feasibility, were averaged per case across reviewers. Paired within-case comparisons between the SFT and GRPO models were assessed using two-tailed paired \textit{t}-tests. For dimensions where normality could not be assumed. All statistical tests were two-sided with a significance threshold of \textit{p} < 0.05. To account for multiple hypothesis testing across evaluation dimensions, \textit{p}-values were adjusted using the Benjamini–Hochberg false discovery rate (FDR) procedure. All analyses were conducted in Python 3.11 using pandas, NumPy, and SciPy. The paired human-evaluation sample consisted of \textit{n} = 100 cases.

\section{Discussion}\label{sec12}

Our findings reveal a striking divergence between algorithmic alignment and clinical alignment in domain-specific medical LLMs.Automatic accuracy across key answer fields indicated that GRPO consistently wined SFT, physician blinded evaluations revealed the opposite pattern. The responses from SFT were rated significantly higher in reasoning quality and treatment feasibility, with no significant differences in diagnostic accuracy. These results demonstrate that benchmark gains do not necessarily translate into higher clinical trust, and highlight a fundamental limitation of current post-training alignment paradigms. Though this divergence exists, both SFT and GRPO were still preferred over the human-level responses in the blinded comparison, illustrating the potential of LLMs as reliable building blocks for clinical decision-support systems. 

We hypothesize that this reverse pattern between superior benchmark scores and superior clinical ratings can be traced to the intrinsic design of the reward function and the optimization dynamics of reinforcement learning. GRPO’s policy is optimized using an answer-centric reward in our implementation, which encourages the model to converge rapidly toward the correct final output with the highest reward scores. This pattern to some extent neglects the process-level reasoning structure that leads to the final answer. As a result, the model tends to produce statistically correct yet clinically plausible answers, which means precise in final answers yet often lacking the stepwise justification physicians expect. Because the  intermediate logical consistency is neither penalized nor rewarded during the alignment, it may drift as a byproduct of the reward updating. The evidence of no significant differences in diagnostic accuracy between two models can also empower the illustration since the initial diagnosis is not covered by the custom reward function in GRPO. In contrast, the SFT model directly mimics the expert natural narrative flow and reasoning steps revealed in the training data. The intention of imitation aligns more closely with physicians’ thinking style, leading to higher scores in clinical ratings. Furthermore, reinforcement learning can induce reward overoptimization, where the policy becomes narrowly tuned to reward signals and gradually diverges from the broader reasoning patterns established during supervised fine-tuning\cite{gao_scaling_2023,coste_reward_2024}. This distribution drift may explain why treatment plans in GRPO responses are sometimes judged as less feasible or less holistic, even when the sole answer is correct. In addition, GRPO generated fewer hallucinations than SFT, likely because continued reward-driven optimization strengthened its adherence to reference data. Both models also outperformed the human reference in the blind winning-rate comparison, indicating that well-aligned LLMs can sometimes produce decisions viewed as more consistent or reliable than individual clinicians. Nonetheless, SFT achieved the highest winning rate overall, reinforcing that imitation-based alignment remains more clinically aligned. Collectively, this gap underscores why accuracy-oriented benchmarks alone cannot capture true clinical alignment, especially in complex, multi-dimensional decision systems like infertility treatment, where many steps are interdependent and difficult to verify.  Achieving meaningful clinical alignment thus requires balancing performance with interpretability, potentially by incorporating reasoning-oriented reward design or structured feedback from physicians during training. 

Beyond the alignment paradox observed in the doctor-in-the-loop evaluation, the field-level results further clarify how different alignment strategies influence specific components of clinical decision-making. GRPO achieves the best on overall average accuracy and macro F1 scores in terms of the main field: infertility type, COS regimen, and  ART strategy. This superior performance is consistent with its outcome-driven reinforcement mechanism. Its improvements in ART strategy and initial diagnosis suggest that reward-based optimization is particularly effective when the decision space is discrete, well-structured, and strongly tied to final outcome correctness. Separately, four models all achieve an accuracy higher than 90\% on Infertility type classification and have no dramatic difference. This demonstrates that the capability of the basic classification problem has been clearly solved and is hard to improve significantly. However, these gains are not uniform. COS regimen prediction remains challenging for all four models, with none outperforming the SFT baseline. According to the physicians’ experience, the COS regimen is more flexible compared to the ART. The regimen choice may vary depending on the physician preference, hormonal response, and cycle history. Because there may be multiple clinically valid regimens for the same patient, an answer-level reward can inadvertently penalize reasonable alternatives, limiting the effectiveness of RL optimization in this field.  In contrast, ART selection benefits from clearer categorical boundaries, enabling GRPO to form more stable reward gradients. These declines reveal the sensitivity of post-training alignment when dealing with complex or ambiguous bounded decisions. The initial diagnosis results further reinforce this pattern. GRPO achieves the highest partial and exact match rates, whereas DPO collapses dramatically. This instability is consistent with preference-learning dynamics: long-form reasoning dilutes the token-level preference signal, causing DPO to converge to over-simplified dominant patterns and lose distributional diversity. By comparison, SFT and ICL preserve more narrative structure and maintain moderate diagnosis performance, although neither improves substantially. Finally, the MAE for Gn dose remains similar across models, indicating that numerical prediction is less sensitive to alignment strategy and likely driven by training data distribution rather than post-training optimization. Together, these results show that reward-driven reinforcement alignment excels in well-defined, outcome-anchored tasks but struggles in flexible or multi-solution domains, whereas SFT preserves reasoning diversity and narrative fidelity at the cost of slightly lower token-level accuracy. These complementary strengths and weaknesses help explain both the doctor-evaluated preference for SFT and the metric-level superiority of GRPO.

A fine-grained breakdown across eleven ART subtypes showed that GRPO substantially improves performance in rare or complex categories (e.g., PGT-M, PGT-SR, Short-protocol IVF), while moderately decreasing performance in ICSI-related cases. This suggests that reinforcement alignment reinforces categories with consistent and well-separated reward signals, but struggles in clinically ambiguous or overlapping decision regions. ICSI lies between IVF (high frequency) and PGT (rare but distinct), receiving neither strong statistical reinforcement nor concentrated reward signals. The reward boundaries for ICSI are inherently fuzzier, limiting gradient stability during optimization. The failure to improve hybrid IVF+ICSI cases further supports the notion that when subtype definitions are clinically mixed, reward-based learning cannot clearly separate decision manifolds. Thus, GRPO amplifies structured decision pathways while dampening mid-range, ambiguous ones, an insight crucial for designing future medical reinforcement learning objectives. Nonetheless, these subtype-level findings align closely with the design of our pyramid-curated alignment dataset. The top “human-enhancement’’ layer and the mid-layer “confusion-enhancement’’ samples were intentionally enriched with rare, complex, and clinically ambiguous scenarios—particularly PGT-M, PGT-SR, short-protocol IVF, and other non-dominant regimens. This targeted data distribution provided GRPO with concentrated, high-signal reward guidance in precisely those long-tail regions, enabling the model to internalize clearer reasoning boundaries and progressively correct SFT’s systematic errors. In contrast, ICSI and its related subtypes are positioned in the middle and bottom of the pyramid hierarchy and characterized by overlapping clinical indications. In this case, these subgroups received weaker and diluted reward signals. As a result, GRPO optimized strongly on top- and mid-layer rare cases but did not achieve stable reinforcement for the more ambiguous ICSI spectrum. This explains the asymmetric improvement pattern observed across subtypes and highlights the importance of hierarchical data construction when aligning medical LLMs for fine-grained decision reasoning.

Both DPO and ICL exhibited inherent weaknesses that limit their usefulness in infertility treatment, or complex clinical reasoning. DPO's pairwise preference optimization depends on token-level contrasts between preferred and rejected answers. The key signals are diluted in long-form reasoning, by a large portion of identical texts between pairs. This leads to instability and collapse in diagnosis-related tasks. ICL improves fluency but lacks adaptive correction mechanisms, making it insufficient for multi-step clinical planning. Together, these limitations highlight that preference- or prompt-based alignment alone cannot support reliable clinical reasoning, underscoring the need for alignment objectives that explicitly integrate both outcome correctness and reasoning fidelity.

Despite its strengths, this study has several technical and methodological limitations that warrant careful interpretation. First, the model’s reasoning dataset relies partially on ICL-generated CoTs rather than fully human-annotated rationales. Although our boutique prompt was curated by specialists, and its generating reliability was proved to be better than normal prompt, the resulting CoTs may encode stylistic biases or propagate inaccuracies from the teacher model. This could affect the fidelity of downstream alignment, particularly in diagnosis-related tasks that require nuanced causal reasoning. Second, while our pyramid dataset effectively amplifies long-tail and confusing cases, its construction is still bounded by the error landscape of the SFT model. If the SFT model exhibits systematic blind spots (e.g., under-specification of certain COS regimens or mid-frequency ICSI cases), these blind spots may propagate into the mid-layer “confusion enhancement” sampling, leading to biased reward allocation during RL alignment. In addition, the EHRs were all collected from a single institution. Although it provides service to a huge population, it could contain demographic and institutional bias. Other than the data constraints, our evaluation focuses on small open-source medical LLMs due to compute constraints. Smaller models are more sensitive to misaligned reward gradients and may underrepresent scaling effects, particularly whether larger models would exhibit the same alignment paradox between benchmark accuracy and clinician trust. The doctor-in-the-loop evaluation, though triple-blind and domain-expert–driven, includes 100 cases. While statistically representative of the ART distribution, this size limits subgroup-level analyses (e.g., rare ICSI variants, short protocol IVF), and may underpower the evaluation of subtle reasoning differences, though we conduct effect sizes computation. Lastly, ground-truth diagnoses and treatment plans represent “single-center ground truth,” which reflects one institution’s practice style. Variability in stimulation preferences, dose selection philosophies, or PGT indications across centers may reduce the generalizability of the aligned model.

Future work should prioritize developing clinically grounded reward engineering, especially by incorporating structured ART guidelines to penalize inconsistent reasoning steps, unsupported causal jumps, and deviations from established treatment pathways, thereby reinforcing fidelity in multi-step clinical thought processes. To improve generalizability, future studies should adopt multi-center and multi-style learning, expanding beyond a single institution to capture diverse treatment philosophies and quantify inter-center variability that may shape ART practice. Another critical direction is conducting scaling experiments with mid-size and large medical LLMs ( > 30B parameters) to determine whether the alignment paradox persists at larger scales, as bigger models may encode richer clinical priors and become less sensitive to reward-induced distribution drift. Additionally, improving reasoning supervision will require human-validated CoT distillation, where a small but expert-curated set of physician-reviewed reasoning chains anchors the logical space and is augmented through LLM-driven self-improvement or targeted red-teaming. Future evaluation frameworks should also move beyond accuracy to include fine-grained clinical trade-offs, such as stimulation risk profiles, embryo yield expectations, dosing safety margins, and adherence to OHSS-prevention heuristics, allowing a more operationally meaningful assessment of model decisions. Finally, given the rapid evolution of ART norms, future work should examine temporal generalization and policy-shift detection, assessing whether aligned models can recognize and adapt to changes in clinical practice—such as shifting COS preferences, increasing PGT utilization, or evolving dosing philosophies—and whether alignment amplifies or suppresses outdated patterns.

\section{Conclusion}\label{sec13}

In this work, we provide the first systematic examination of how post-training alignment strategies—SFT, DPO, GRPO, and ICL—shape the clinical behavior of medical LLMs in a real-world, multi-stage decision environment. Across both automatic metrics and triple-blind physician evaluations, our results reveal a consistent pattern: reinforcement-based optimization improves outcome-level correctness but does not necessarily enhance, and may even erode, clinician-perceived reasoning quality and treatment feasibility. This divergence, which we term the \textit{alignment paradox}, highlights a structural gap between algorithmic optimization and clinical trust.

Our fine-grained analyses further show that GRPO reallocates learning capacity along the ART hierarchy: it strengthens reasoning in complex and long-tail subgroups, particularly PGT and short-protocol IVF, yet underperforms in mid-frequency, clinically ambiguous categories such as ICSI. These dynamics reflect the sensitivity of reward-based alignment to data distribution and reward granularity, and underscore the need for carefully designed supervisory signals in medical domains.

Through the introduction of a pyramid-curated dataset and a dual-evaluation framework, we demonstrate that both SFT and GRPO can outperform human practitioners in case-level decision consistency, but excel in fundamentally different ways, SFT through reasoning transparency and GRPO through outcome-driven precision. Taken together, our findings emphasize that building clinically reliable medical LLMs requires more than maximizing benchmark accuracy: it demands alignment strategies that integrate structured clinical knowledge, preserve reasoning faithfulness, and center the evaluative criteria of physicians. As the field advances toward real-world deployment, bridging this gap between algorithmic alignment and clinical alignment will be essential for developing trustworthy, interpretable, and safe AI systems in reproductive medicine and beyond.

\backmatter
 \newpage  
\section*{Declarations}

\subsection{Funding}
Part of this study was supported funded by the Science and Technology Department of Sichuan Province Project (2024YFFK0365); Part of this study was supported funded by the Natural Science Foundation of Sichuan, China (2025NSFSC1985); Part of this study was supported by the 1·3·5 project for disciplines of excellence, West China Hospital, Sichuan University (ZYYC21004).  

\subsection{Conflict of interest}
None declare

\subsection{Ethics approval}
A set of selected masked EHRs recorded between 2020 and 2022 in the Infertility Outpatient Department at West China Second University Hospital was considered in this study. The study was approved by the Ethics Committee of West China Second University Hospital, Sichuan University (ID: 2022288)

\subsection{Data availability }
The datasets generated during and/or analyzed during this study are not publicly available due to institutional case privacy and a large number of interaction dialogs, but are available from the corresponding author on reasonable request. The authors will make the Author Accepted Manuscript (AAM) version available under a CC BY public copyright license.

 \newpage 
  % 或 sn-basic / sn-apa / sn-mathphys 等
\bibliography{references}      % 或 sn-bibliography.bib

\end{document}